%% file: main.tex
\documentclass[10pt,twocolumn,letterpaper]{article}

\usepackage{cvpr}              
\usepackage{multirow}
\usepackage{booktabs}
\usepackage{graphicx}
\usepackage{amssymb}
\usepackage{fontawesome5}
\usepackage{xcolor}
\definecolor{mygold}{HTML}{FFD700}
\input{preamble}
\definecolor{cvprblue}{rgb}{0.21,0.49,0.74}
\usepackage[pagebackref,breaklinks,colorlinks,allcolors=cvprblue]{hyperref}

\def\paperID{925} 
\def\confName{CVPR}
\def\confYear{2026}

\title{MMTIT-Bench: A Multilingual and Multi-Scenario Benchmark with Cognition–Perception–Reasoning Guided Text-Image Machine Translation}

\author{
Gengluo Li$^{1,4}$ \quad
Chengquan Zhang$^{2,\ddagger}$ \quad
Yupu Liang$^{5}$ \quad
Huawen Shen$^{1,4}$ \quad
Yaping Zhang$^{5}$ \\
Pengyuan Lyu$^{2}$ \quad
Weinong Wang$^{2}$ \quad
Xingyu Wan$^{2}$ \quad
Gangyan Zeng$^{6}$ \\
Han Hu$^{2}$ \quad
Can Ma$^{1,4}$\textsuperscript{\scalebox{0.8}{\faEnvelope}} \quad
Yu Zhou$^{3}$\textsuperscript{\scalebox{0.8}{\faEnvelope}} \\[0.25em]
$^{1}$Institute of Information Engineering, Chinese Academy of Sciences \quad
$^{2}$Tencent \\
$^{3}$Nankai University \quad
$^{4}$University of Chinese Academy of Sciences \\
$^{5}$Institute of Automation, Chinese Academy of Sciences \\
$^{6}$Nanjing University of Science and Technology \\[0.25em]
{\tt\small ligengluo@iie.ac.cn \quad yzhou@nankai.edu.cn \quad macan@iie.ac.cn} \\
{\small $^{\ddagger}$ Project leader \quad
\textsuperscript{\scalebox{0.9}{\faEnvelope}} Corresponding author}
}

\begin{document}
\maketitle
\input{sec/0_abstract}    
\input{sec/1_intro}
\input{sec/2_related_work}

\input{sec/3_benchmark}

\input{sec/4_cascade_data}
\input{sec/5_ex}
\input{sec/6_conclusion}
\input{sec/acknowledge}
{
    \small
    \bibliographystyle{ieeenat_fullname}
    \bibliography{main}
}


\end{document}

%% file: sec/0_abstract.tex
\begin{abstract}
End-to-end text-image machine translation (TIMT), which directly translates textual content in images across languages, is crucial for real-world multilingual scene understanding. Despite advances in vision–language large models (VLLMs), robustness across diverse visual scenes and low-resource languages remains underexplored due to limited evaluation resources. We present \textbf{MMTIT-Bench}, a human-verified multilingual and multi-scenario benchmark with 1,400 images spanning fourteen non-English and non-Chinese languages and diverse settings such as documents, scenes, and web images, enabling rigorous assessment of end-to-end TIMT. Beyond benchmarking, we study how reasoning-oriented data design improves translation. Although recent VLLMs have begun to incorporate long Chain-of-Thought (CoT) reasoning, effective thinking paradigms for TIMT are still immature: existing designs either cascade parsing and translation in a sequential manner or focus on language-only reasoning, overlooking the visual cognition central to VLLMs. We propose \textbf{Cognition–Perception–Reasoning for Translation (CPR-Trans)}, a data paradigm that integrates scene cognition, text perception, and translation reasoning within a unified reasoning process. Using a VLLM-driven data generation pipeline, CPR-Trans provides structured, interpretable supervision that aligns perception with reasoning. Experiments on 3B and 7B models show consistent gains in accuracy and interpretability. We will release \textbf{MMTIT-Bench} to promote the multilingual and multi-scenario TIMT research upon acceptance.
\end{abstract}

%% file: sec/1_intro.tex
\section{Introduction}
\label{sec:intro}
\begin{figure}[ht]
    \centering
    \includegraphics[width=1\linewidth]{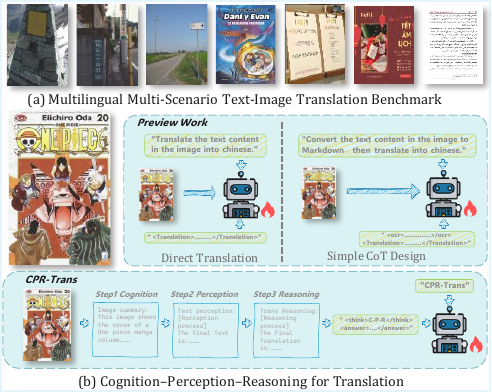}
    \vspace{-14pt}
    \caption{MMTIT-Bench and the CPR-Trans Data Paradigm.}
    \vspace{-10pt}
    \label{fig:intro}
\end{figure}

Document and scene text understanding underpin a wide range of real-world applications, including information retrieval, content digitization, and multilingual communication~\cite{rag_doc,rag_storm}.
Among these tasks, \textbf{Text-Image Machine Translation (TIMT)}~\cite{ma2024born, Dota}, which directly translates textual content in images across languages, has emerged as a key capability for real-world multilingual intelligence.

Early approaches~\cite{intro_cascade, zhang2023novel} mainly follow a \textit{cascade paradigm}, where optical character recognition (OCR) is first applied to extract text then a separate module performs translation.
While modular and interpretable, this pipeline suffers from error propagation, weak visual grounding, and limited robustness in complex layouts.

With the rapid advancement of vision–language large models (VLLMs)~\cite{mllm1,Qwen2.5-VL}, research has shifted toward \textit{end-to-end TIMT}, which maps visual inputs to target-language text within a unified framework.
Although recent studies report strong results under \textbf{digital-born or synthetic conditions}~\cite{mllm_got,mllm_vary}, where text is clean and well-aligned and languages are mostly high-resource such as English and Chinese, realistic multilingual translation across diverse visual domains remains underexplored.
This limitation largely stems from the lack of comprehensive benchmarks that evaluate VLLMs across multiple languages and real-world scenes.
In addition, the design of an effective \textbf{reasoning-oriented data paradigm} to guide VLLMs in TIMT has not been sufficiently investigated.

To address these issues, we introduce the \textbf{Multilingual Multi-Scenario Text-Image Machine Translation Benchmark (MMTIT-Bench)}, a human-verified benchmark for end-to-end TIMT across fourteen non-English and non-Chinese languages (Figure~\ref{fig:intro}(a)).
Each image is annotated with bilingual translations in Chinese and English, enabling fine-grained cross-lingual evaluation under realistic visual conditions.
MMTIT-Bench contains 1{,}400 carefully curated samples that cover diverse scenarios such as menus, posters, books, products, and attractions, providing a comprehensive and challenging testbed for assessing TIMT systems.

While recent multimodal models have begun integrating long Chain-of-Thought (CoT) reasoning to enhance visual understanding~\cite{cot1,cot2,cot3}, a \textbf{thinking paradigm} tailored for TIMT remains an open problem.
Existing attempts~\cite{tans_reason, ocrtrans} are often cascaded designs that separate parsing and translation, or language-only CoT formulations that overlook the visual cognition central to VLLMs.
To this end, we propose \textbf{Cognition–Perception–Reasoning for Translation (CPR-Trans)}, a native reasoning-oriented data paradigm that tightly integrates visual scene cognition, text perception, and translation reasoning within a unified multimodal process (Figure~\ref{fig:intro}(b)).
Through a VLLM-driven data generation pipeline, CPR-Trans provides structured and interpretable supervision that aligns perception and reasoning, ultimately enhancing end-to-end translation capability.
Comprehensive experiments on both \textbf{3B} and \textbf{7B} model scales show that CPR-Trans substantially improves translation accuracy and interpretability, highlighting the value of reasoning-oriented data design in advancing TIMT.

\noindent \textbf{To sum up, our contributions are threefold:}
\begin{itemize}
    \item \textbf{MMTIT-Bench.}
    We present a multilingual and multi-scenario benchmark for TIMT covering 14 non-English and non-Chinese languages, addressing the lack of evaluation resources under diverse real-world conditions.
    \item \textbf{Comprehensive evaluation of VLLMs.}
    We systematically evaluate a range of open-source and closed-source VLLMs to assess their multilingual and multi-scenario translation capabilities.
    \item \textbf{CPR-Trans paradigm.}
    We propose a unified reasoning-oriented data paradigm that models \textit{cognition}, \textit{perception}, and \textit{reasoning} within the translation process.
    Experiments on both 3B and 7B models show consistent gains in translation accuracy and interpretability.
\end{itemize}

%% file: sec/2_related_work.tex
\section{Related Work}
\label{sec:formatting}


\noindent\textbf{Text-Image Machine Translation.}
TIMT aims to translate textual content embedded in images from a source language into a target language~\cite{TIMT_intro1,TIMT_intro2,Dota,team2025hunyuanocr}.
Existing approaches mainly follow two paradigms.
\textit{Cascade methods}~\cite{Casecade2} perform Optical Character Recognition (OCR) followed by Neural Machine Translation (NMT), but suffer from error propagation, weak visual grounding, and inefficiency in complex or noisy scenes.
In contrast, \textit{end-to-end methods}~\cite{E2E1,E2E2,liang2025single} jointly model visual encoding and text decoding within a unified architecture, while recent VLLMs further enhance translation accuracy.


\begin{table*}[ht]
\centering
\renewcommand{\arraystretch}{1.03}
\small
\caption{Comparison of existing Text-Machine Translation benchmarks.}
\vspace{-10pt}
\setlength{\aboverulesep}{0ex}
\setlength{\belowrulesep}{0ex}
\renewcommand{\arraystretch}{1.1}
\begin{tabular}{lcccccccc}
\toprule
\multirow{2}{*}{\textbf{Dataset}}  & \textbf{Sample}  & \textbf{Avg. Words} & \multirow{2}{*}{\textbf{Languages}} & \multicolumn{3}{c}{\textbf{Scenarios}} & \multicolumn{2}{c}{\textbf{Annotation}} \\ \cmidrule(lr){5-7} \cmidrule(lr){8-9}
 & \textbf{Nums} & \textbf{Per Image} & & Web & Document & Street & Model & Human \\ \midrule
PATIMT~\cite{patimt} & 1200 & 507 & En$\to$Zh, Zh$\to$En & \checkmark & & & & \checkmark \\
DITrans~\cite{1DITrans,ditrans2} & 1651 & 458 & En$\to$Zh & & \checkmark & & & \checkmark \\
DoTA~\cite{Dota} & 1003 & 447 & En$\to$Zh & & \checkmark & & \checkmark & \checkmark \\
OCRMT~\cite{OCRMT} & 1000 & 5 & Zh$\to$En & & & \checkmark & \checkmark & \checkmark \\ 
MTIT6~\cite{mtit6}  &  1200  &  7  &  (En\&Ko\&Ja)$\to$Zh, Zh$\to$(En\&Ko\&Ja)  &  \checkmark  &  &  \checkmark  &  \checkmark  &  \checkmark \\ \midrule
\textbf{MMTIT (Ours)} & 1400 & 160 & 14 Langs$\to$(En \& Zh) & \checkmark & \checkmark & \checkmark & \checkmark & \checkmark \\ \bottomrule
\end{tabular}
\vspace{-10pt}
\label{benchmark_comparison}
\end{table*}


\noindent\textbf{TIMT Benchmarks.}
Existing TIMT benchmarks remain limited in language coverage and visual diversity.
Early datasets such as DITrans~\cite{1DITrans,ditrans2} and DoTA~\cite{Dota} mainly focus on English–Chinese document translation with human or semi-automatic annotations, while OCRMT~\cite{OCRMT} targets street-view scenes and PATIMT~\cite{patimt} provides synthetic samples for controlled evaluation.
MTIT6~\cite{mtit6} explores real-world scenarios but covers only four languages.
Despite supporting early research, these datasets have limited linguistic scope and image diversity, restricting comprehensive evaluation and fair assessment of model generalization in real-world settings.

\noindent\textbf{Reasoning-Oriented data paradigms for TIMT.}
In end-to-end TIMT, the design of data paradigms is pivotal for model performance. Prior work~\cite{ocrtrans} mitigated the degradation of OCR perception by concatenating OCR outputs with translation labels, yielding clear translation gains. Recent advances on Long-CoT supervision~\cite{openaio1,liu2024deepseek,yang2025qwen3} show that decomposing, verifying, and refining intermediate steps can substantially enhance complex reasoning. However, reasoning-oriented supervision tailored to TIMT remains in its infancy: existing approaches often rely on cascaded OCR to NMT pipelines or apply language-only CoT that overlooks visual cognition. The potential of a native, multimodally grounded Long-CoT data paradigm to improve end-to-end TIMT thus remains largely unexplored.

%% file: sec/3_benchmark.tex
\section{Multilingual Multi-Scenario Text-Image Translation Benchmark}

To address the lack of comprehensive multilingual and multi-domain benchmarks for end-to-end TIMT, we construct the \textbf{MMTIT-Bench}, designed to capture diverse visual scenes and lingual variations, providing a realistic testbed for evaluating VLLMs' translation capabilities.

\begin{figure*}[ht]
    \centering
    \includegraphics[width=1\linewidth]{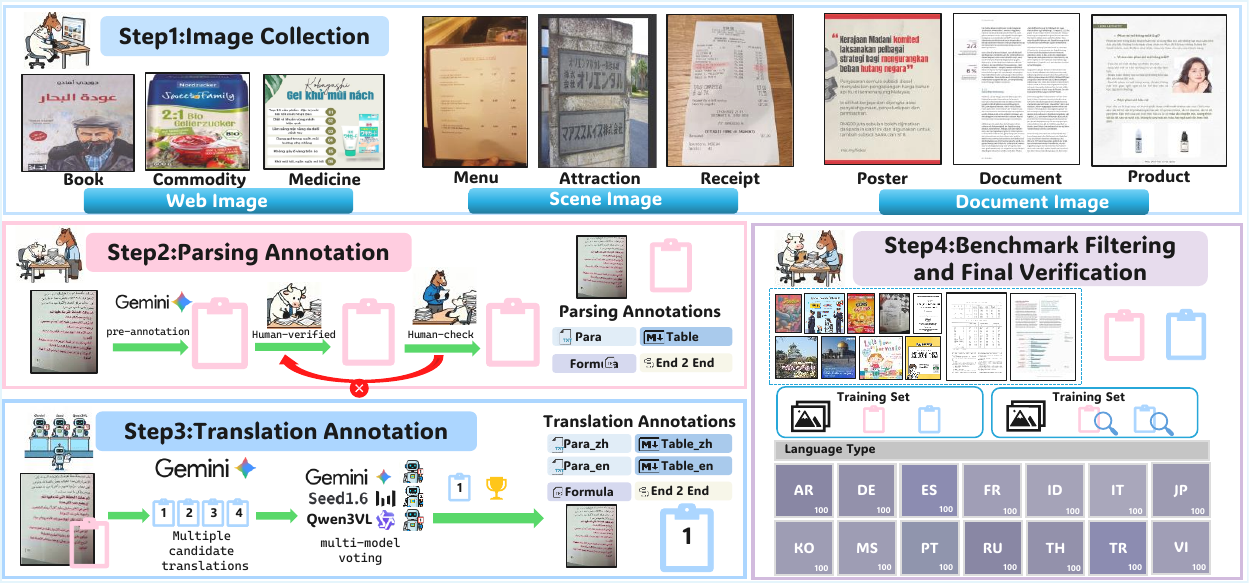}
    \vspace{-15pt}
    \caption{Overview of the \textbf{MMTIT-Bench} construction pipeline, which consists of four stages: image collection, parsing annotation, translation annotation, and benchmark filtering with human verification.}
    \vspace{-10pt}
    \label{fig:shoutu}
\end{figure*}

Compared with existing TIMT benchmarks that focus on limited language settings or single-domain scenarios, MMTIT-Bench offers broader lingual coverage and visual diversity. It includes fourteen non-English and non-Chinese languages across multiple real-world scenarios, such as documents, web, and scene images. Built through a model-assisted generation and human post-editing pipeline, MMTIT-Bench strikes a balance between scalability and annotation quality, enabling reliable evaluation under realistic multilingual translation conditions. A detailed comparison with existing benchmarks is summarized in Table~\ref{benchmark_comparison}, with representative examples and the annotation workflow shown in Figure~\ref{fig:shoutu}. We next detail the image collection and annotation pipeline behind MMTIT-Bench.

\subsection{Image Collection}

To capture the diversity of real-world multimodal translation scenarios, we collect images across 14 high-frequency non-English and non-Chinese languages, including German, Spanish, Turkish, Vietnamese, Korean, Malay, Portuguese, Russian, French, Indonesian, Thai, Italian, and Japanese.
All images are manually gathered by annotators through UI-based search channels to ensure authenticity and accurate language representation.

The collection covers a wide range of diverse scenarios such as attractions, menus, posters, books, documents and receipts, enabling comprehensive evaluation under diverse lingual and visual conditions.
For each language, approximately 1,000 images containing naturally occurring text are collected and manually verified for text clarity, language correctness, and balanced domain coverage, yielding around 14,000 high-quality samples for subsequent annotation and evaluation.

\subsection{MLLM-Assisted Data Annotation}
\label{subsec:annotation}

The annotation of MMTIT-Bench involves two major stages: \textit{text parsing annotation} and \textit{translation annotation}, combining MLLM-assisted generation with human post-editing to ensure both scalability and quality.

\subsubsection{Text Parsing Annotation}
\label{subsubsec:text_parsing}

For each collected image, we first perform OCR-based pre-annotation using \textit{Gemini~2.5~Flash}, which provides initial text recognition results. Human annotators then carefully revise these outputs, correcting misrecognized characters and reorganizing the text to follow the \textit{natural reading order} within the image. During this process, tabular contents are normalized into standard Markdown format, and mathematical expressions are reformatted into \LaTeX{} syntax to ensure structural clarity and consistency.

To guarantee annotation reliability, all parsed texts undergo mutual cross-checking among annotators. Each sample is independently reviewed by another annotator to verify transcription accuracy, reading order correctness, and overall consistency with the visual content.

\subsubsection{Translation Annotation}
\label{subsubsec:translation_annotation}

In the translation stage, we adopt an automated VLLM-based annotation pipeline. Each image, along with its verified OCR text, is fed into \textit{Gemini~2.5~Flash}, which is prompted to translate all recognized text within the image in a single pass. Multi-temperature sampling is applied to generate multiple candidate translations for each sample.

To enhance reliability, we employ a multi-model voting mechanism, where \textit{Gemini~2.5~Flash}, \textit{Seed~1.6~Thinking}, and \textit{Qwen3-VL~Thinking} serve as independent evaluators. Each model assesses all candidate translations and votes for the most accurate and natural result. The translation with the highest number of votes is selected as the final annotation; in case of a tie, one of the top-voted candidates is randomly chosen.

This process yields two parallel translation sets—one translating from the source language to Chinese and the other to English—providing comprehensive bidirectional coverage for evaluating text-image translation quality within the benchmark.

\subsection{Benchmark Filtering and Final Verification}
\label{subsec:benchmark_filtering}

After completing annotation, we conduct a human-in-the-loop filtering stage to ensure the overall quality and balance of the benchmark. For each of languages, we select 100 images based on the richness and representativeness of their visual scenes, resulting in \textbf{1,400 fully annotated samples}. All selected samples then undergo a final round of manual verification and correction, conducted by language experts for the respective languages, covering both parsed texts and bilingual translations (Chinese and English).

This curated subset constitutes the official evaluation split of \textbf{MMTIT-Bench}, providing high-quality, human-refined references for assessing end-to-end TIMT. The remaining annotated data, excluding these benchmark samples, are included in the \textbf{training corpus} used in subsequent experiments to investigate the impact of different reasoning-oriented data construction paradigms.

\subsection{Evaluation Protocol}
\label{subsec:evaluation_protocol}

We evaluate translation quality under two complementary protocols: a VLLM-based judging paradigm and a rule-based metric protocol. This dual setting enables both human-aligned semantic assessment and conventional score comparability.

\noindent\textbf{(1) VLLM-based Judging.}
Motivated by evidence that large language models align well with human judgments in translation evaluation~\cite{huang2024lost,guo2025automatic}, we adopt two representative VLLM judges: the closed-source \textit{Gemini~2.5~Flash} and the open-source \textit{Qwen3-VL-235B-A22B-Instruct}. 
Each judge receives the original image, the reference translation, and the system output, then assigns four scores—\textit{faithfulness}, \textit{fluency}, \textit{readability}, and \textit{terminology consistency}. 
The overall score is the mean of these four dimensions. 
For fairness, only the final translation text is evaluated for thinking-enabled models, excluding intermediate reasoning traces. 
Complete prompts and the scoring rubric are provided in the supplementary material.

\noindent\textbf{(2) Rule-based Metrics.}
To complement VLLM-based evaluation, we also report results using the standard \textit{COMET} metric~\cite{comet}, a widely adopted automatic translation quality measure based on reference-based regression modeling. This conventional evaluation provides a quantitative baseline to compare against semantic judgments from multimodal VLLM judges.


\subsection{Benchmark Evaluation Experiments}
\label{subsec:evaluation_experiments}

\begin{table*}[ht]
\centering
\small
\caption{Comparison of translation performance across different VLLMs under various evaluation settings. The “think” column indicates whether the model employs explicit reasoning.}
\vspace{-10pt}
\renewcommand{\arraystretch}{1.2}
\setlength{\tabcolsep}{5pt}
\setlength{\aboverulesep}{0ex}
\setlength{\belowrulesep}{0ex}
\begin{tabular}{llcccccccc}
\toprule
\multirow{2}{*}{\textbf{\#}} & \multirow{2}{*}{\textbf{Model}} & \multirow{2}{*}{\textbf{Param.}} & \multirow{2}{*}{\textbf{Think}} &
\multicolumn{2}{c}{\textbf{Gemini-flash Judge}} &
\multicolumn{2}{c}{\textbf{Qwen3VL Judge}} &
\multicolumn{2}{c}{\textbf{COMET}} \\
\cmidrule(lr){5-6} \cmidrule(lr){7-8} \cmidrule(lr){9-10}
& & & & other2en & other2zh & other2en & other2zh & other2en & other2zh \\
\midrule

\multirow{2}{*}{\parbox{1.2cm}{\textbf{Cascade\\Pipeline}}} & MinerU2.5\cite{mineru25}+Qwen3   &  -   &  - & 48.32 & 49.70 & 45.13 & 46.26 & 63.43 & 65.06\\
& Dots.ocr\cite{dotsvllm}+Qwen3   &  -   &  - & 66.24 & 65.27 & 61.20 & 60.98 & 73.62 & 73.42 \\ \midrule
\multirow{7}{*}{\parbox{1.2cm}{\textbf{General\\VLMs}}} & MiMo-VL\cite{Mimo} & 7B & \checkmark & 67.72 & 69.52 & 64.81 & 65.24 & 74.96 & 76.15 \\
& dots.vlm1\cite{dotsvllm} & 671B-A37B & \checkmark & \underline{79.23} & 76.73 & \underline{73.82} & 72.14 & \underline{76.84} & 78.45 \\
& Qwen3-VL-Instruct\cite{Qwen2.5-VL} & 235B-A22B & - & 64.39 & 69.67 & 61.75 & 65.00 & 73.67 & 77.20 \\
& Qwen3-VL-Thinking\cite{Qwen2.5-VL} & 235B-A22B & \checkmark & 73.81 & \underline{77.90} & 70.38 & 72.14 & 76.21 & \underline{78.81} \\
& Seed1.6 & - & - & 67.21 & 66.35 & 62.61 & 65.45 & 74.01 & 76.55 \\
& Seed1.6-Thinking & - & \checkmark & 74.09 & 75.82 & 73.72 & \underline{73.90} & 76.23 & 78.77 \\
& Gemini2.5-flash\cite{gemini25} \textcolor{mygold}{\faMedal} & - & \checkmark & \textbf{82.94} & \textbf{85.00} & \textbf{74.76} & \textbf{76.26} & \textbf{79.26} & \textbf{80.06} \\
\midrule
\midrule
\multirow{3}{*}{\parbox{1.2cm}{\textbf{Ours}}}
& Qwen2.5-VL$_{\mathrm{Origin}}$ & 7B & - & 53.98 & 46.89 & 52.03 & 47.06 & 70.27 & 71.67 \\
& Qwen2.5-VL$_{\mathrm{sft(Direct)}}$ & 7B & - & 68.40 & 62.42 & 69.04 & 66.37 & 75.42 & 76.81 \\
& Qwen2.5-VL$_{\mathrm{sft(CPR\text{-}Trans)}}$ & 7B & \checkmark & \textbf{83.98} & \underline{82.84} & \textbf{82.99} & \textbf{82.23} & \textbf{81.07} & \textbf{81.04} \\
\bottomrule
\end{tabular}
\label{tab:judge_comparison}
\vspace{-10pt}
\end{table*}

We evaluate a series of representative VLLMs, including both open-source and proprietary systems, on the proposed \textbf{MMTIT-Bench}. All models are tested in an end-to-end Text-Image Machine Translation setting under identical prompting conditions for fairness. For comparison, we additionally include a \textit{Cascade Baseline} built upon an OCR–LLM pipeline~\cite{mineru25,dots.ocr2025,yang2025qwen3}, where a strong text parsing model first extracts textual content from images and the \textit{Qwen3-Instruct-235B-A22B} model performs translation.

\noindent\textbf{Evaluation Results.}
As summarized in Table~\ref{tab:judge_comparison}, both VLLM-based judging and rule-based metric protocols yield highly consistent scoring trends, demonstrating strong alignment between human-aligned and metric-based evaluations. 
Among all systems, \textit{Gemini~2.5-Flash} achieves the best overall performance, reflecting superior generalization and robustness in end-to-end TIMT.  
In contrast, the \textit{Cascade Baseline} shows inferior performance, mainly due to the error propagation inherent to OCR-dependent pipelines, which struggle with visually complex or non-digital-born scenes.  
Within the same model families (e.g., \textit{Qwen3-VL} and \textit{Seed1.6}), the thinking-mode variants consistently outperform non-thinking versions, confirming that explicit reasoning modeling effectively enhances translation capability.

Beyond zero-shot evaluation, we further fine-tune \textit{Qwen2.5-VL-7B} using our proposed \textbf{CPR-Trans} data paradigm, which introduces structured reasoning supervision tailored for translation.  
This fine-tuning yields competitive performance compared with models of substantially larger parameter scales, highlighting the effectiveness of reasoning-oriented data construction for improving TIMT.
The following section details the design principles and methodology behind the \textbf{CPR-Trans} paradigm.

%% file: sec/4_cascade_data.tex
\section{Designing Effective Chain-of-Thought Paradigms for TIMT}
\label{sec:paradigm}

Recent studies have shown that incorporating \textbf{CoT} reasoning significantly enhances model performance across complex reasoning tasks. 
This effect extends to VLLMs, where enabling explicit reasoning leads to consistent gains on multimodal understanding and translation benchmarks. 
Our evaluation also confirms that these improvements in TIMT arise not from model scale, but from the design of reasoning-oriented data paradigms. 
Thus, the central question becomes: \textit{what kind of CoT design can most effectively enhance translation performance in VLLMs?}

Existing TIMT data paradigms can be broadly categorized into three types.  
1) \textit{Direct Translation} provides only the final translation text, often causing models to lose OCR perception and multimodal grounding.  
2) \textit{Simple CoT Design} augments translation with explicit OCR reproduction, preserving text recognition but neglecting essential reasoning stages such as scene cognition, text perception and translation reasoning.  
3) \textit{Native Thinking in VLLMs} allows the model to perform self-directed reasoning during translation; however, this process is uncontrolled, often producing redundant or cyclic reflections that lead to unstable translation quality.

To address these limitations, we propose a cognitively inspired data construction framework, \textbf{Cognition–Perception–Reasoning for Translation (CPR-Trans)}, which explicitly structures the reasoning process for TIMT.  
CPR-Trans aligns with the natural human translation process:the model first performs \textit{scene cognition} to capture holistic information understanding,
then conducts \textit{text perception} to identify and interpret textual regions, and finally engages in \textit{translation reasoning} to generate faithful and contextually coherent translations.  
This structured paradigm provides interpretable and controllable reasoning supervision that bridges perception and generation, effectively enhancing end-to-end translation capability.

\label{subsec:cpr_pipeline}

\begin{figure*}[ht]
    \centering
    \includegraphics[width=1\linewidth]{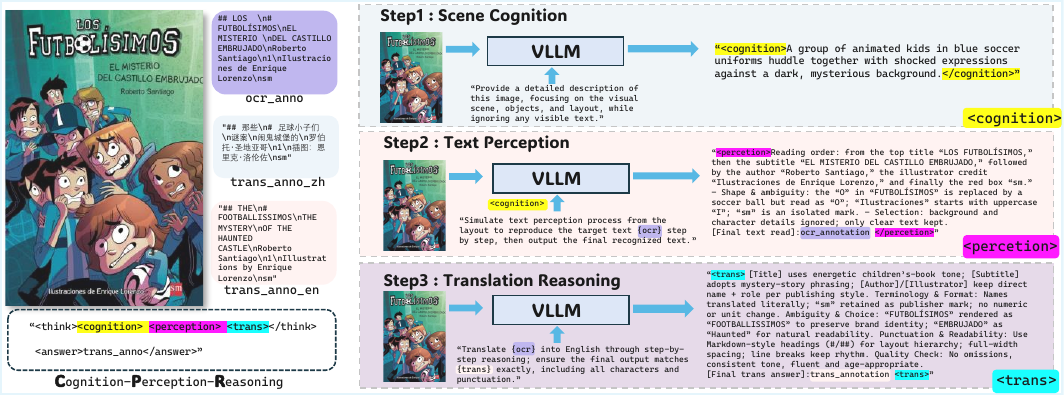}
    \vspace{-15pt}
    \caption{Overview of the CPR-Trans data generation pipeline.}
    \vspace{-10pt}
    \label{fig:casecadepipeline}
\end{figure*}

\noindent\textbf{CPR-Trans Data Generation Pipeline.} To construct training data aligned with the proposed \textit{CPR-Trans} paradigm, we design a \textit{VLLM-assisted multi-stage generation pipeline}, as illustrated in Figure~\ref{fig:casecadepipeline}. 
The pipeline is implemented using \textit{Qwen3-VL-235B-A22B-Instruct} and guided by structured prompts to produce reasoning traces corresponding to each cognitive stage of the translation process.

\begin{itemize}
    \item \textbf{Step 1 – Scene Cognition.}  
    Given an input image, the model describes the global visual context—objects, layout, and background—while explicitly excluding any text recognition behavior.  
    The output of this stage forms the visual understanding caption, denoted as \texttt{<cognition>}.
    
    \item \textbf{Step 2 – Text Perception.}  
    The model then receives the image, human OCR annotations, and the \texttt{<cognition>} result.  
    It reasons about the spatial arrangement, reading order, and structure of textual regions, producing an organized perception description consistent with human OCR results, denoted as \texttt{<perception>}.
    
    \item \textbf{Step 3 – Translation Reasoning.}  
    Using the image, OCR annotations, and human translations, the model explains the translation reasoning process, integrating visual and textual understanding to derive the final translation output, denoted as \texttt{<trans>}.
\end{itemize}

\noindent\textbf{Integrated Construction.}  
The three components are concatenated into a unified \textbf{CPR reasoning sequence}, explicitly encoding the full translation reasoning chain.  
To distinguish reasoning from the final translation, the entire reasoning trace is enclosed within \texttt{<think></think>} tags, and the ground-truth translation within \texttt{<answer></answer>}.  
This structured annotation provides explicit reasoning supervision and a well-defined translation target for TIMT model training.  
Detailed prompt templates and implementation details are provided in the supplementary material.

%% file: sec/5_ex.tex
\section{Experiments}
\begin{table*}[t]
\centering
\renewcommand{\arraystretch}{1.03} %
\setlength{\tabcolsep}{11.5pt}
\setlength{\aboverulesep}{0ex}
\setlength{\belowrulesep}{0ex}
\small
\caption{Comparison of different data paradigms under 3B and 7B model settings, evaluated by Gemini-Flash, Qwen3VL, and COMET.}
\vspace{-10pt}

\begin{tabular}{clcccccc}
\toprule
\multirow{2}{*}{\parbox{1.2cm}{\textbf{Model\\Param.}}} & \multirow{2}{*}{\ \ \ \ \ \ \ \ \textbf{Method}} &
\multicolumn{2}{c}{\textbf{Gemini-Flash Judge}} & 
\multicolumn{2}{c}{\textbf{Qwen3VL Judge}} &
\multicolumn{2}{c}{\textbf{COMET}} \\ \cmidrule(l){3-4} \cmidrule(l){5-6} \cmidrule(l){7-8}
&  & other2en & other2zh & other2en & other2zh & other2en & other2zh \\ 
\midrule
\multirow{6}{*}{\textbf{3B}} 
& Origin & 38.32 & 33.04 & 47.01 & 43.54 & 68.80 & 69.96 \\
& Direct & 64.98 & 58.83 & 66.67 & 63.86 & 73.98 & 74.29 \\
& Simple CoT & \underline{72.31} & \underline{67.18} & \underline{74.06} & \underline{71.51} & \underline{77.62} & \underline{78.21} \\
& Distillation(VLLM) & 69.15 & 66.76 & 68.01 & 66.12 & 75.38 & 76.12 \\
& Distillation(LLM) & 67.92 & 65.22 & 67.30 & 64.51 & 75.21 & 75.09 \\ \cmidrule(l){2-8}
& \textbf{CPR-Trans} & \textbf{82.00} & \textbf{81.29} & \textbf{81.39} & \textbf{81.32} & \textbf{80.50} & \textbf{80.32} \\ \midrule
\multirow{6}{*}{\textbf{7B}} 
& Origin & 53.98 & 46.89 & 52.03 & 47.09 & 70.27 & 71.67 \\
& Direct & 68.40 & 62.42 & 69.04 & 66.37 & 75.42 & 76.81 \\
& Simple CoT & \underline{74.65} & \underline{71.03} & \underline{76.30} & \underline{73.00} & \underline{79.69} & \underline{79.24} \\
& Distillation(VLLM) & 71.90 & 69.91 & 69.09 & 67.51 & 76.93 & 77.83 \\
& Distillation(LLM) & 69.37 & 65.39 & 69.23 & 67.21 & 75.82 & 76.90 \\ \cmidrule(l){2-8}
& \textbf{CPR-Trans} & \textbf{83.98} & \textbf{82.84} & \textbf{82.99} & \textbf{82.23} & \textbf{81.07} & \textbf{81.04} \\
\bottomrule
\end{tabular}
\label{tab:cascade_results}
\end{table*}

\begin{table*}[t]
\centering
\small
\caption{Ablation study of reasoning components on 3B and 7B models under two evaluation settings. }
\vspace{-10pt}
\renewcommand{\arraystretch}{0.99} %
\begin{tabular}{cccccccccccc}
\toprule
\multirow{2}{*}{\textbf{\#}} & \multicolumn{3}{c}{\textbf{Component}} & \multicolumn{2}{c}{\textbf{Gemini2.5 (o2e)}} & \multicolumn{2}{c}{\textbf{Gemini2.5 (o2z)}} 
& \multicolumn{2}{c}{\textbf{Qwen3VL (o2e)}} & \multicolumn{2}{c}{\textbf{Qwen3VL (o2z)}} \\ \cmidrule(l){2-4} \cmidrule(l){5-6} \cmidrule(l){7-8} \cmidrule(l){9-10} \cmidrule(l){11-12}
& Cognition & Perception & Trans & 3B & 7B & 3B & 7B & 3B & 7B & 3B & 7B \\
\midrule
Baseline &  &  &  & 72.31 & 74.65 & 67.18 & 71.03 & 74.06 & 76.30 & 71.51 & 73.00 \\ \midrule
1 & \checkmark &  &  & 75.23 & 76.91 & 72.21 & 74.12 & 76.82 & 79.08 & 72.88 & 75.69 \\
2 &  & \checkmark &  & 72.40 & 74.43 & 67.88 & 71.21 & 73.53 & 75.69 & 71.56 & 72.36 \\
3 &  &  & \checkmark & 79.47 & 80.73 & 78.92 & 80.21 & 78.49 & 80.51 & 76.92 & 80.32 \\ \midrule
4 & \checkmark & \checkmark &  & 76.98 & 77.82 & 73.90 & 75.50 & 77.03 & 79.73 & 73.45 & 77.26 \\
5 & \checkmark &  & \checkmark & 80.01 & \underline{82.11} & 79.45 & \underline{81.32} & 78.99 & 81.39 & 77.56 & \underline{81.20} \\
6 &  & \checkmark & \checkmark & \underline{81.29} & 81.90 & \underline{80.09} & 81.20 & \underline{80.87} & \underline{82.30} & \underline{80.45} & 81.11 \\ \midrule
\textbf{Final (Full)} & \checkmark & \checkmark & \checkmark & \textbf{82.00} & \textbf{83.98} & \textbf{81.29} & \textbf{82.84} & \textbf{81.39} & \textbf{82.99} & \textbf{81.32} & \textbf{82.23} \\
\bottomrule
\end{tabular}
\label{tab:ablation_results}
\vspace{-10pt}
\end{table*}

\subsection{Experimental Setup}

\paragraph{Training Data Design.}
Our training corpus includes both manually annotated and synthesized data.  
The manual subset is derived from the annotated pool not used for building \textbf{MMTIT-Bench}, covering 12{,}600 \textit{Other→Chinese} and 12{,}600 \textit{Other→English} samples with verified OCR and translation annotations.  

To scale up, we synthesize 70{,}000 additional image–text pairs across the same 14 languages using \textit{SynthDog}~\cite{synthdog}.  
Texts are drawn from open translation corpora and validated by \textit{Qwen3-235B-A22B-Instruct} for accuracy and consistency.  
In total, the final corpus contains \textit{165{,}200} aligned multimodal samples, each paired with OCR annotations and bilingual translation labels.

\noindent\textbf{Comparative Data Paradigms.}
To examine how different data construction paradigms affect end-to-end TIMT performance, we prepare four training variants as follows:

\vspace{-3pt}
\begin{itemize}
    \item \textbf{Direct Translation.}  
    Each sample includes only the final translation label, without any OCR text or reasoning.  

    \item \textbf{Simple CoT Design.}  
The verified OCR result is directly concatenated with the corresponding translation output within a single training sample, forming a simple \texttt{[OCR Parsing + Translation]} annotation without explicit reasoning traces.

    \item \textbf{Thinking-Model Distillation.}  
    Reasoning-augmented data are distilled from thinking-mode teachers.  
    For the VLLM-based distillation, \textit{Qwen3-VL-235B-A22B-Thinking} receives raw images and produces multimodal reasoning traces,  
    while for the LLM-based distillation, \textit{Qwen3-235B-A22B-Thinking} takes human OCR annotations as textual input to generate reasoning-augmented translations.  
    All generated outputs are verified for semantic consistency with human references using \textit{Gemini~2.5~Flash}, resulting in \textit{145,721} VLLM-distilled and \textit{152,021} LLM-distilled samples.  

    \item \textbf{CPR-Trans.}  
    Following the CPR structure, each sample integrates three reasoning stages into a unified \texttt{<think>...</think><answer>...</answer>} sequence.  
\end{itemize}
\vspace{-3pt}

\noindent\textbf{Base Models and Fine-tuning Configuration.}
For fair comparison across model scales, we adopt \textit{Qwen2.5-VL-7B} and \textit{Qwen2.5-VL-3B}
This setup allows us to assess both the generalization and scalability of different data paradigms.  
All models are fine-tuned for 3 epochs with full-parameter optimization at a learning rate of $3\times10^{-5}$ on 32 NVIDIA H20 GPUs, and results are reported from the final checkpoint on MMTIT-Bench.

\subsection{Comparison of Data Paradigms}

The comparison results of the four data construction paradigms are summarized in Table~\ref{tab:cascade_results}.

Overall, the \textit{Direct Translation} paradigm yields the weakest performance across both model scales.  
Lacking explicit perception or reasoning supervision, models in this setting often overfit to shallow text–image correlations and gradually lose their OCR perception ability, resulting in lower translation accuracy.
In contrast, the \textit{Simple CoT Design} incorporates explicit OCR parsing supervision and achieves clear improvements, demonstrating that maintaining OCR comprehension effectively preserves visual–text grounding and enhances translation fluency.

The \textit{Thinking-Model Distillation} paradigm, distilled from both language-only and vision–language reasoning sources, improves over direct translation but remains inferior to the Simple CoT setting.
Distillation from a VLLM-based teacher consistently yields better translation than from an LLM-based one, suggesting that reasoning grounded in visual cognition provides more effective signals for TIMT.
However, both types of distilled data share similar issues: the reasoning traces generated by thinking-mode models are often uncontrolled, redundant, or noisy.
Under limited data supervision, such unstable reasoning sequences add unnecessary complexity to learning and hinder the model from internalizing structured reasoning patterns.

\begin{figure*}[ht]
    \centering
    \includegraphics[width=1\linewidth]{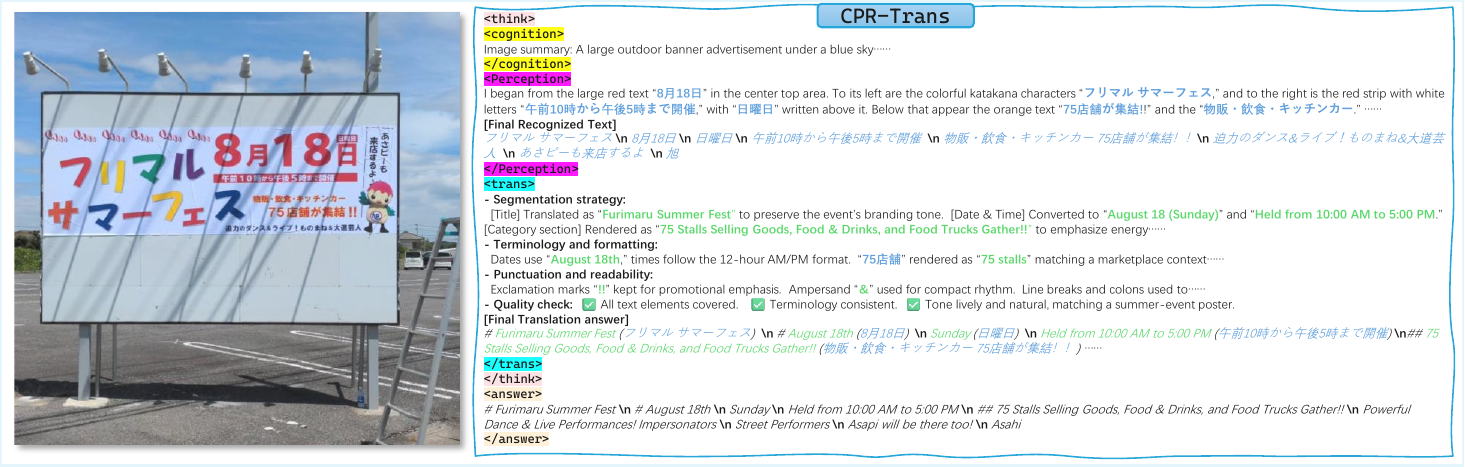}
    \vspace{-15pt}
    \caption{Qualitative example of CPR-Trans on an MMTIT-Bench sample, showing how the model sequentially conducts scene cognition, text perception, and translation reasoning to produce the final translation.}
    \vspace{-10pt}
    \label{fig:leida}
\end{figure*}

Notably, our proposed \textit{CPR-Trans paradigm} achieves the best results across all model settings.
Unlike distillation-based reasoning, which passively inherits the teacher’s reasoning noise, CPR-Trans introduces a well-defined, cognitively grounded reasoning structure that explicitly models \textit{cognition}, \textit{perception}, and \textit{reasoning} within a unified framework.  
This disciplined reasoning process allows the model to exploit the advantages of chain-of-thought supervision in a controllable and interpretable manner, leading to consistent and significant gains in translation accuracy and semantic fidelity across model scales.


\subsection{Ablation Study on CPR-Trans Components}

To examine the contribution of each reasoning component within \textbf{CPR-Trans}, we conduct a systematic ablation study covering all combinations of its three stages: \textit{Cognition}, \textit{Perception}, and \textit{Reasoning}.  
Each model variant selectively removes or replaces one or more components while keeping the overall training setup identical, enabling a fine-grained analysis of how each stage and their interactions influence translation performance.  
Specifically, the cognition stage is excluded to test the effect of global visual understanding, the perception reasoning is replaced with plain OCR annotations to assess the role of spatial and structural reasoning beyond raw recognition, and the translation reasoning is replaced with direct translation labels to compare purely supervised translation against reasoning-guided generation.

Results are summarized in Table~\ref{tab:ablation_results}.  
Across both \textbf{3B} and \textbf{7B} model scales, and under evaluations from \textit{Gemini~2.5-Flash} and \textit{Qwen3-VL}, consistent patterns emerge.  
Removing cognition causes a clear performance drop, showing that grounding translation in holistic visual context helps the model make more accurate lexical and semantic choices.  
Perception reasoning alone contributes limited improvement, yet when combined with cognition or translation reasoning, it produces additional gains, indicating that perception primarily acts as a bridge that enhances multimodal consistency.  
Translation reasoning yields the most substantial improvement, confirming that explicit modeling of linguistic transformation and decision-making directly improves both faithfulness and fluency.

Overall, the complete \textbf{CPR-Trans} configuration achieves the best performance across all settings.  
Compared with the baseline using only OCR and translation labels, it yields average gains of \textbf{+11.2} under \textit{Gemini~2.5-Flash} and \textbf{+8.2} under \textit{Qwen3-VL}.  
These results demonstrate that cognition, perception, and reasoning provide complementary benefits, and their integration forms a coherent, stepwise reasoning chain that significantly enhances end-to-end multimodal translation performance.

\begin{table}[t]
\centering
\small
\caption{Training-free Validation of the CPR-Trans Paradigm on Qwen3-VL-235B-A22B-Instruct.}
\vspace{-10pt}
\renewcommand{\arraystretch}{1.03} %
\setlength{\aboverulesep}{0.5ex}
\setlength{\belowrulesep}{0.5ex}
\begin{tabular}{ccccc}
\toprule
\multirow{2}{*}{\parbox{1.2cm}{\textbf{Inference\\Strategy}}} & \multicolumn{2}{c}{\textbf{Gemini2.5 Judge}} & \multicolumn{2}{c}{\textbf{Qwen3VL Judge}} \\
\cmidrule(l){2-3} \cmidrule(l){4-5}
& other2en & other2zh & other2en & other2zh \\
\midrule
\textbf{One-step} & 64.39 & 69.67 & 61.75 & 65.00 \\
\textbf{Multi-step} & \textbf{70.54} & \textbf{73.21} & \textbf{68.72} & \textbf{70.54} \\
\bottomrule
\end{tabular}
\label{tab:training_free}
\vspace{-10pt}
\end{table}

\subsection{Training-free Validation of the CPR-Trans}

To further validate the intrinsic effectiveness of our paradigm without additional fine-tuning, we conduct a training-free reasoning experiment using \textbf{Qwen3-VL-235B-A22B-Instruct}.
We compare two inference strategies:
1) a \textbf{single-turn} setting, where the model directly performs end-to-end translation in one round of interaction; and
2) a \textbf{multi-turn} setting, where the model is guided to follow the \textbf{CPR-Trans} process across three dialogue turns—first describing the scene, then analyzing text and layout, and finally generating the translation.
Results are shown in Table~\ref{tab:training_free}.

The \textbf{multi-turn} setting consistently improves translation quality across both evaluation protocols.
This shows that guiding the model to follow the cognition–perception–reasoning process through multi-turn interaction enhances translation accuracy and coherence even without additional training, confirming that \textbf{CPR-Trans} naturally aligns with the cognitive mechanism of TIMT.

\subsection{Qualitative Visualization on MMTIT-Bench}

To illustrate the effectiveness of our approach, we visualize representative samples from \textbf{MMTIT-Bench} using the \textit{Qwen2.5-VL-7B} model fine-tuned under the \textbf{CPR-Trans} paradigm.
As shown in Figure~\ref{fig:leida}, the model exhibits a clear and interpretable reasoning flow that integrates scene cognition, text perception, and translation generation, producing accurate and contextually consistent results.
These visualizations demonstrate the robustness and transparency of CPR-Trans in guiding TIMT reasoning.
More examples are provided in the supplementary material.


%% file: sec/6_conclusion.tex
\section{Conclusion}

We present \textbf{MMTIT-Bench}, a multilingual and multi-scenario benchmark that offers a comprehensive evaluation framework for TIMT across fourteen diverse languages. We further propose \textbf{CPR-Trans}, a reasoning-oriented paradigm that unifies cognition, perception, and reasoning within the translation process. Extensive experiments across different model scales demonstrate that CPR-Trans substantially improves translation accuracy, coherence, and interpretability. We believe that the proposed benchmark and paradigm provide a solid foundation for advancing cognitively grounded TIMT, and can further inspire future extensions toward video and document-level translation tasks.

%% file: sec/acknowledge.tex
\section*{Acknowledgements} 

This work is supported by the National Natural Science Foundation of China (Grant NO 62376266 and 62406318).